\documentclass[10pt,twocolumn,letterpaper]{article}

\usepackage[pagenumbers]{wacv}      % To produce the REVIEW version for the applications track
\newcommand{\parnum}[1]{\textbf{(#1)}}
\definecolor{wacvblue}{rgb}{0.21,0.49,0.74}
\usepackage[pagebackref,breaklinks,colorlinks,allcolors=wacvblue]{hyperref}

\def\wacvPaperID{2425} % *** Enter the WACV Paper ID here
\def\confName{WACV}
\def\confYear{2027}

\title{FlexMap: Robust HD Map Construction under Flexible Camera Configurations}

\author{First Author\\
Institution1\\
Institution1 address\\
{\tt\small firstauthor@i1.org}
\and
Second Author\\
Institution2\\
First line of institution2 address\\
{\tt\small secondauthor@i2.org}
}

\author{
Run Wang\textsuperscript{1} \quad
Chaoyi Zhou\textsuperscript{1} \quad
Amir Salarpour\textsuperscript{1} \quad
Xi Liu\textsuperscript{1} \quad
Zhi-Qi Cheng\textsuperscript{2}\\
Feng Luo\textsuperscript{1} \quad
Mert D. Pesé\textsuperscript{1}%
\thanks{Corresponding author: \texttt{mpese@clemson.edu}.}
\quad
Siyu Huang\textsuperscript{1}\\[4pt]
\textsuperscript{1}School of Computing, Clemson University\\
\textsuperscript{2}Tacoma School of Engineering and Technology,
University of Washington
}

\begin{document}
\maketitle

\begin{abstract}
High-definition (HD) maps provide essential semantic information about road structures for autonomous driving, but existing HD map construction methods typically require calibrated multi-camera rigs and explicit 2D-to-BEV transformations. Such pipelines degrade when camera views are missing or pose estimates are inaccurate, limiting their use across heterogeneous fleet configurations. We introduce FlexMap, a vectorized HD mapping framework that adapts to varying camera configurations without architectural changes or per-configuration retraining and does not require camera parameters as model input. FlexMap replaces explicit geometric projection with a geometry foundation model that encodes cross-view 3D structure. A spatial-temporal enhancement module then separates cross-view spatial reasoning from temporal aggregation, while a camera-aware decoder uses the token produced for each input view to adapt its attention without camera poses. Experiments on nuScenes and Argoverse 2 show that FlexMap outperforms pose-dependent baselines using estimated poses and maintains comparable accuracy across all evaluated camera configurations, including those with missing views.

\end{abstract}

\section{Introduction}

\begin{figure}[!t]
    \centering
    \includegraphics[width=1\linewidth]{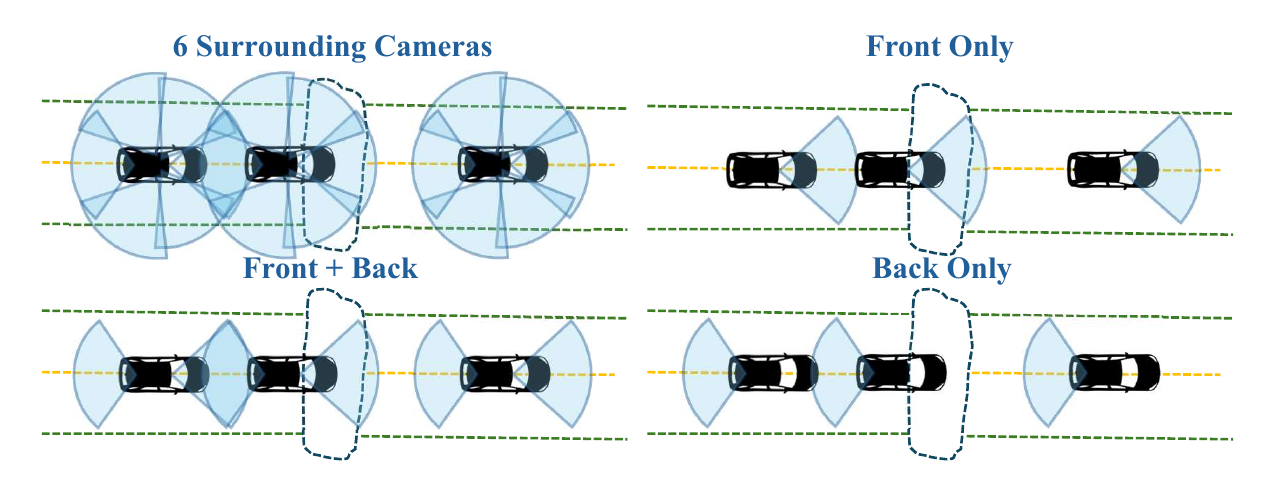}
    \caption{
    % \textbf{FlexMap} enables HD map construction and remains strong under diverse camera setups.
    FlexMap uses one shared model to construct HD maps from single-, paired-, and surround-view camera configurations.
    }
    \label{fig:teaser}
\end{figure}

High-definition (HD) maps provide essential semantic information about road structures for autonomous driving systems~\citep{vazquez_deep_2022, scheel_urban_2021, jo_simultaneous_2018, du_rtmap_2025, zhang_mapnav_2025, HDMapSurvey, ding_pivotnet_2023}. Recent end-to-end vectorized mapping approaches~\citep{maptr2023, streammapnet, himap2024} enable real-time map generation from onboard sensors, but scaling them to crowdsourced deployment remains challenging. Real-world vehicle fleets are heterogeneous (\cref{fig:teaser}), with different camera configurations and view availability. A general HD mapping system must therefore operate robustly across varying camera configurations with one shared model.

\begin{figure}[!b]
    \centering
    \includegraphics[width=1\linewidth]{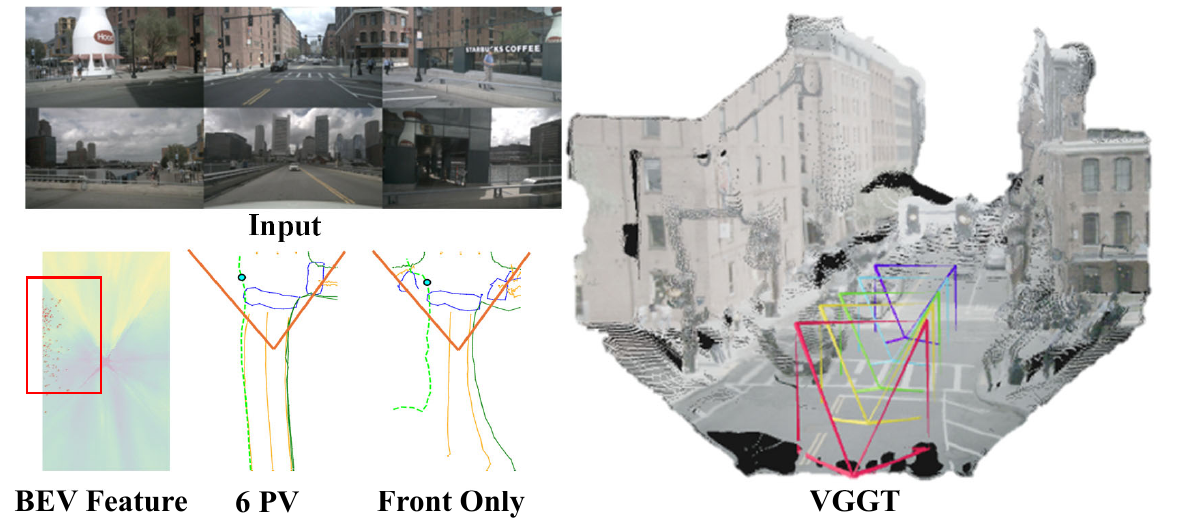}
    \caption{
    We visualize the BEV feature map used by a BEV-based decoder (e.g., MapTR~\citep{maptr2023}) and a query (green) whose attention for a target location concentrates near a view boundary (red box). The full six-camera setting (6PV) yields coherent vectors; removing the other views eliminates their cross-view contributions and produces fragmentation around the same boundary. VGGT retains coherent cross-view structure for a sequence of front-only images, motivating its use in a flexible HD map pipeline.}
    \label{fig:camera_missing_results}
\end{figure}

To support robust and cost-effective deployment, we study the construction of vectorized HD maps from images without camera parameters as input. Representative approaches such as MapTR~\citep{maptr2023} and GeMap~\citep{gemap2024} rely on explicit 2D-to-BEV (bird's-eye-view) projection modules~\citep{liu_bevfusion_2024,chen_efficient_2022} that require camera intrinsics and extrinsics. These pose-dependent pipelines degrade when views are missing or pose estimates are inaccurate~\citep{hao_safemap_2025,hao_what_2025, hao_is_2024}. For example, removing a contributing camera eliminates cross-view support near the corresponding BEV boundary and fragments the decoded vectors (\cref{fig:camera_missing_results}).
\label{sec:decoder}
To decouple vectorized HD mapping from explicit camera pose, we build on recent feed-forward geometry foundation models (GFMs)~\citep{wang_vggt_2025}, which infer dense cross-view 3D structure without camera parameters. Adapting these models to driving scenes raises two challenges: (1) preserving static road structures across time under transient occlusion, and (2) inferring spatial relationships without an explicit 2D-to-BEV projection.

To address these challenges, we introduce \textbf{FlexMap}, a feed-forward architecture for flexible camera configurations. FlexMap replaces explicit geometric projection with a pre-trained geometry backbone that produces cross-view, geometry-grounded features. A spatial-temporal enhancement module separately models cross-view and cross-time context, while a camera-aware decoder uses the token produced for each input view to adapt query initialization and attention. Together, these components construct vectorized HD maps across camera configurations without requiring camera parameters as model input.

Our contributions are summarized as follows:
\begin{enumerate}
    \item We formulate vectorized HD map construction from flexible camera configurations without using camera parameters as model input.
    \item To the best of our knowledge, FlexMap is the first vectorized HD map construction framework to use a geometry foundation model in place of explicit 2D-to-BEV projection and decode camera-centric vector maps without camera parameters during map prediction.
    \item Experiments on nuScenes~\citep{caesar_nuscenes_2020} and Argoverse 2~\citep{wilson_argoverse_2023} show that FlexMap outperforms pose-dependent baselines using estimated poses across the evaluated camera configurations, including settings with missing views.
\end{enumerate}

\section{Related work}
We review prior work on (i) vectorized HD map construction, (ii) robust mapping with incomplete or corrupted camera inputs, and (iii) GFMs that provide geometry-aware representations for cross-view reasoning.

\noindent\textbf{Vectorized HD map construction.}
Early methods formulated mapping as BEV semantic segmentation~\citep{chen_efficient_2022, zhou_cross-view_2022, hu_fiery_2021, li_bevformer_2022}, whereas recent approaches directly predict structured polylines for downstream planning~\citep{li2021hdmapnet, liu_vectormapnet_2023}. MapTR and related query-based methods~\citep{maptr2023,maptrv22024,himap2024} established the common paradigm of parallel decoding with permutation-equivalent queries. Later work improves temporal consistency~\citep{maptrv22024,streammapnet}, introduces geometric priors~\citep{gemap2024}, or refines query interactions~\citep{hu_admap_2024, wang_stream_2024, kim_unveiling_2024, interactionmap2025}. In each case, multi-view image features are first aligned in a shared BEV frame before vector decoding. These architectures therefore rely on pose-dependent 2D-to-BEV transformations built around a predefined camera setup, which limits their use with changing camera configurations.

\noindent\textbf{Robust HD mapping.}
Most mapping frameworks are designed and evaluated with the fully calibrated multi-camera rigs defined by standard benchmarks~\citep{caesar_nuscenes_2020, wilson_argoverse_2023}. In real driving systems, sensors can be corrupted, unavailable, or miscalibrated~\citep{dong_benchmarking_2023, zhu_understanding_2023, kong_robodrive_2024, ge_metabev_2023}. Recent evaluations~\citep{hao_safemap_2025} show large performance drops under incomplete or noisy inputs. A key source of this sensitivity is the use of calibration-dependent view transformations~\citep{li_bevformer_2022, liu_bevfusion_2024, chen_efficient_2022}. FlexMap instead maps each view without camera intrinsics or extrinsics.

\noindent\textbf{Geometry foundation models.}
Recent feed-forward geometry models~\citep{wang_vggt_2025,wang_dust3r_2024,keetha_mapanything_2026}, such as VGGT~\citep{wang_vggt_2025}, extract cross-view 3D structure without camera parameters as input. Their geometry-aware features support 3D reconstruction and novel view synthesis~\citep{jiang_anysplat_2025, zuo_dvgt_2025, lin_vgd_2025}, but have not been developed for vectorized HD mapping. These features alone do not provide the semantic classes, ordered point sets, or camera-centric outputs required by the mapping task. Applying them to HD mapping therefore requires converting implicit 3D geometry into semantically labeled, ordered camera-centric BEV polylines while preserving flexibility across view directions and camera counts.

\section{Methodology}
\begin{figure*}[!htbp]
    \centering
    \includegraphics[width=\linewidth]{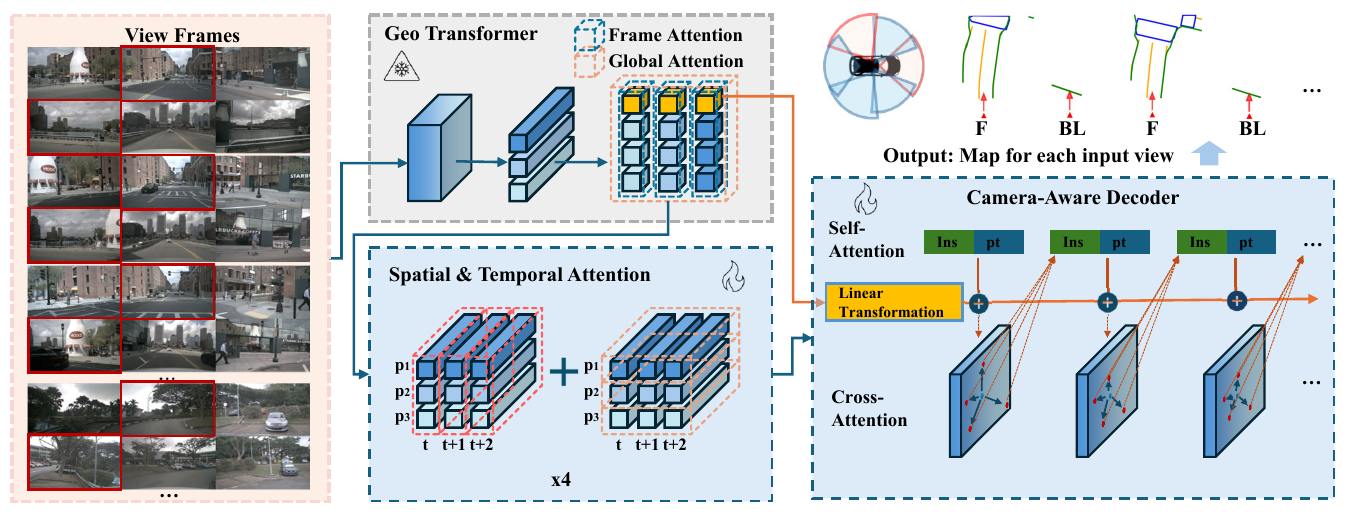}
    \caption{\textbf{Overview of our method.} Multi-camera temporal sequences are processed by a geometry transformer to extract multi-scale features. A spatial-temporal attention module aggregates information across views and time, and a camera-aware decoder with hierarchical queries predicts a vectorized map for each input view. The example uses front and back-left camera inputs.}
    \label{fig:pipeline}
\end{figure*}

We propose a feed-forward pipeline for high-definition map construction from temporal multi-camera sequences. As shown in \cref{fig:pipeline}, it contains three stages: geometry feature extraction and fusion, spatial-temporal enhancement, and camera-aware vector decoding. FlexMap first extracts and fuses geometry-grounded features, then enhances them across views and time, and finally predicts camera-centric vectorized map elements for every input view.

\subsection{Problem setup}
Consider a configuration of $V$ cameras observed over $T$ timesteps, giving $S = V \times T$ frames $\{\mathbf{I}_s\}_{s=1}^{S}$, where $\mathbf{I}_s \in \mathbb{R}^{H \times W \times 3}$. FlexMap receives no camera intrinsics or extrinsics as model input. For each input view $s$, it predicts $N$ vectorized map elements $\{(\mathbf{v}_i^s, c_i^s)\}_{i=1}^{N}$ in a camera-centric BEV frame. Here, $\mathbf{v}_i^s \in \mathbb{R}^{P \times 2}$ is an ordered polyline of $P$ points with $(x,y)$ coordinates, and $c_i^s$ is its class label (divider, pedestrian crossing, or boundary).
% \subsection{Problem Setup}
% Consider $S$ uncalibrated input views $\{\mathbf{I}_s\}_{s=1}^{S}$, where each image $\mathbf{I}_s \in \mathbb{R}^{H \times W \times 3}$ has height $H$, width $W$, and 3 color channels. For each view, we predict $N$ vectorized map elements $\{(\mathbf{v}_i^s, c_i^s)\}_{i=1}^{N}$, where $\mathbf{v}_i \in \mathbb{R}^{P \times 2}$ is a polyline  (an ordered sequence of $P$ points with x, y coordinates.), and $c_i$ is the class label indicating the type of map element (e.g., lane divider, pedestrian crossing, or road boundary).

\subsection{Geometry transformer}
\label{sec:feature_extraction}
% \noindent\textbf{Geometry Transformer.} 
Following VGGT~\citep{wang_vggt_2025}, we patchify and encode each image into DINOv2 visual tokens~\citep{oquab_dinov2_2024}, with one camera token prepended to each input view. The geometry transformer uses alternating attention: even layers apply frame attention within each view to model its spatial structure, while odd layers apply global attention across views and timesteps. After this process, the camera token $\mathbf{t}_s$ summarizes the visual content and cross-view geometry of view $s$, while the visual tokens encode spatial, temporal, and cross-view context for downstream map decoding.

\noindent\textbf{Feature fusion.} 
We extract intermediate features from multiple layers and fuse them. These features are projected to a common channel dimension, reshaped spatially, and progressively fused through residual convolutions, yielding perspective-view (PV) features $\mathbf{F}_s \in \mathbb{R}^{H' \times W' \times C}$, where $C$ denotes the shared feature dimension.

\subsection{Spatial-temporal enhancement}
\label{sec:spatiotemporal}
Spatial relations across cameras and temporal relations across frames require different aggregation patterns. We therefore separate cross-view spatial attention from cross-time temporal attention. Let $\nu_s$ denote the camera stream of frame $s$ and $\tau_s$ its time index. These values are used only to form attention groups for two operations:

\noindent\textbf{Cross-view attention.}
Frames sharing a timestamp ($\tau_i = \tau_j$) exchange information across camera views. This operation models cross-view spatial structure and can connect observations of the same lane from the front and front-left cameras.

\noindent\textbf{Temporal attention.}
Frames from the same camera ($\nu_i = \nu_j$) exchange information across timestamps. This operation models the apparent motion of lane markings and the persistence of static road structures. Separating the two operations prevents cross-view and cross-time relationships from being mixed in a single attention operation. The resulting features $\tilde{\mathbf{F}}_s$ remain in perspective-view space while carrying context from neighboring views and timestamps.

\begin{figure*}[!htbp]
    \centering
    \includegraphics[width=\linewidth]{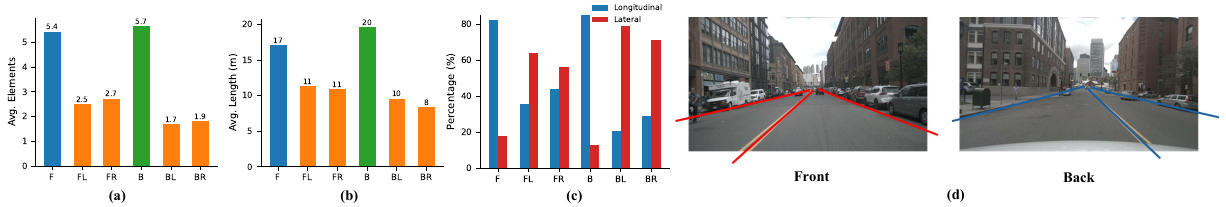}
    \caption{\textbf{Camera-specific map element statistics for the nuScenes~\citep{caesar_nuscenes_2020} dataset after FOV clipping.} (a) Average number of elements, (b) average polyline length, and (c) orientation distribution (longitudinal vs.\ lateral). Front and back cameras (F, B) observe denser, longer structures, while side cameras (FL, FR, BL, BR) capture fewer, shorter, and largely lateral elements. Panel (d) illustrates how camera-centric map geometry differs across cameras: the front view primarily contains longitudinal lane structures aligned with the driving direction, whereas the back view presents the road geometry under the reversed viewing direction.}
    \label{fig:method_camera_aware_combined}
\end{figure*}

% \begin{figure}[!htbp]
%     \centering
%     \includegraphics[width=0.9\linewidth]{Figures/camera_distribution_combined.pdf}
%     \caption{\textbf{Camera-specific map element statistics for nuScenes \citep{caesar_nuscenes_2020} dataset after FOV clipping.}
%     (a) Average number of elements, (b) average polyline length, (c) orientation distribution (longitudinal vs.\ lateral), and (d) class-wise element counts across camera views. Front and back cameras (F, B) observe denser, longer structures, while side cameras (FL, FR, BL, BR) capture fewer, shorter, and largely lateral elements.}
%     \label{fig:camera_distribution}
% \end{figure}

% \begin{figure}[!htbp]
%     \centering
%     \includegraphics[width=\linewidth]{Figures/method_camera_aware.pdf}
%     \caption{\textbf{Qualitative illustration of view-dependent geometry.}
%     Vectorized map elements projected in perspective view differ systematically across cameras. The front view primarily contains longitudinal lane structures aligned with the driving direction, whereas the back view exhibits distinct geometry due to the reversed viewing direction.}    \label{fig:camera_front_back}
% \end{figure}

\subsection{Camera-aware decoder}
\label{sec:decoder-method}
Different cameras exhibit systematic differences in the geometry and density of observable map elements. As quantified in \cref{fig:method_camera_aware_combined}, front and back views contain denser and longer structures with a longitudinal orientation, whereas side views contain fewer, shorter, and more lateral segments. \Cref{fig:method_camera_aware_combined} also provides a qualitative example of how geometry differs by camera. These view-dependent statistics motivate a decoder that conditions both query initialization and attention on the camera token produced for each view.

\noindent\textbf{Hierarchical query design.} Following MapTR~\citep{maptr2023}, we use a hierarchical query structure with $N$ instance embeddings and $P$ point embeddings combined via addition. Each query is represented by $\mathbf{q} \in \mathbb{R}^{C}$.

\noindent\textbf{View-conditioned initialization.} The camera token $\mathbf{t}_s$ encodes the visual content and cross-view geometry of input view $s$. We project it to the query dimension:
\begin{equation}
\mathbf{c}_s = \mathcal{F}_{\text{linear}}(\mathbf{t}_s) \in \mathbb{R}^{C}.
\end{equation}
Initial reference points are generated from both the query and the projected camera token:
\begin{equation}
\mathbf{r}^0 = \sigma\left(\mathcal{F}_{\text{ref}}([\mathbf{q}; \mathbf{c}_s])\right) \in [0,1]^2.
\end{equation}
Conditioning reference-point generation on $\mathbf{c}_s$ allows the decoder to learn view-specific initialization patterns for different camera orientations.

\noindent\textbf{View-adaptive deformable attention.}
Each decoder layer applies self-attention followed by deformable cross-attention~\citep{zhu_deformable_2021} to perspective-view features, where each query samples $K$ locations. For the layer-$l$ query $\mathbf{q}^l$, both sampling offsets and attention logits are conditioned on the camera token:
\begin{equation}
\Delta\mathbf{r}^l = \mathcal{F}_{\text{offset}}([\mathbf{q}^l; \mathbf{c}_s]), \quad
\mathbf{a}^l = \mathcal{F}_{\text{weight}}([\mathbf{q}^l; \mathbf{c}_s]).
\end{equation}
This conditioning allows the decoder to learn view-specific sampling patterns for different camera orientations.

\noindent\textbf{View-conditioned temperature scaling.}
Map elements vary in geometry and density across camera views, motivating view-adaptive attention sharpness. We modulate attention sharpness via a view-conditioned temperature $\tau$ predicted per attention head:
\begin{equation} 
\tau = \sigma(\mathcal{F}_{\tau}(\mathbf{c}_s)) \cdot (\tau_{\max} - \tau_{\min}) + \tau_{\min},
\end{equation}
which scales the attention logits before applying softmax:
\begin{equation}
\hat{\mathbf{a}}^l = \text{softmax}(\mathbf{a}^l / \tau).
\end{equation} 
The deformable cross-attention output is computed by sampling features at offset locations and aggregating them with the normalized attention weights:
\begin{equation}
\mathbf{z}^l = \mathcal{F}_{\text{out}}\left(\sum_{k=1}^{K} \hat{a}_k^l \cdot \mathbf{F}_s(\mathbf{r}^l + \Delta\mathbf{r}_k^l)\right),
\end{equation}
where $\mathbf{F}_s(\cdot)$ bilinearly interpolates features at the sampled locations and $\mathcal{F}_{\text{out}}$ is the output projection. Residual and feed-forward updates produce $\hat{\mathbf{q}}^l$ from $\mathbf{z}^l$. At each layer, we update the reference points using the updated query and camera token:
\begin{equation}
\mathbf{r}^{l+1} = \sigma\left(\mathcal{F}_{\text{ref}}([\hat{\mathbf{q}}^l; \mathbf{c}_s]) + \sigma^{-1}(\mathbf{r}^l)\right).
\end{equation}
The final query embeddings are used to produce map-element predictions through two parallel branches: a classification branch that predicts semantic class probabilities and a polyline regression branch that predicts $P$ normalized 2D coordinates in the camera-centric bird's-eye view.

\subsection{Training objective}
Predictions are matched to ground truth via Hungarian matching with permutation-invariant costs~\citep{maptr2023}. We apply deep supervision across all $L$ decoder layers and all $S$ input views:
\begin{equation}
\mathcal{L} = \frac{1}{SL} \sum_{s=1}^{S} \sum_{l=1}^{L} \left(\lambda_{\text{cls}} \mathcal{L}_{\text{cls}}^{l,s} + \lambda_{\text{pts}} \mathcal{L}_{\text{pts}}^{l,s} + \lambda_{\text{dir}} \mathcal{L}_{\text{dir}}^{l,s}\right)
\end{equation}
where $\mathcal{L}_{\text{cls}}$ is the focal loss for classification, $\mathcal{L}_{\text{pts}}$ is the L1 loss for point coordinates, and $\mathcal{L}_{\text{dir}}$ is the cosine direction loss between predicted and target polylines.

\section{Experiments}
\label{sec:experiments}

\begin{table*}[ht]
\centering
\caption{Comparison on the nuScenes~\citep{caesar_nuscenes_2020} validation set across camera configurations. Baselines use estimated camera poses; their results with ground-truth poses are shown in parentheses. MapTRv2$^\ast$ is retrained using estimated poses. FlexMap uses no pose input.}

\setlength{\tabcolsep}{5.0pt}
\renewcommand{\arraystretch}{1.1}

\resizebox{\textwidth}{!}{%
\begin{tabular}{lcccc|cccc|cccc|cccc}
\toprule
& \multicolumn{4}{c|}{\textbf{Front only (1PV)}}
& \multicolumn{4}{c|}{\textbf{Back only (1PV)}}
& \multicolumn{4}{c|}{\textbf{Front+Back (2PV)}}
& \multicolumn{4}{c}{\textbf{All 6 PVs (6PV)}} \\
\cmidrule(lr){2-5}\cmidrule(lr){6-9}\cmidrule(lr){10-13}\cmidrule(lr){14-17}
\textbf{Method}
& $\text{AP}_{\text{ped}}$ & $\text{AP}_{\text{div}}$ & $\text{AP}_{\text{bou}}$ & \textbf{mAP}
& $\text{AP}_{\text{ped}}$ & $\text{AP}_{\text{div}}$ & $\text{AP}_{\text{bou}}$ & \textbf{mAP}
& $\text{AP}_{\text{ped}}$ & $\text{AP}_{\text{div}}$ & $\text{AP}_{\text{bou}}$ & \textbf{mAP}
& $\text{AP}_{\text{ped}}$ & $\text{AP}_{\text{div}}$ & $\text{AP}_{\text{bou}}$ & \textbf{mAP} \\
\midrule
MapTR~\citep{maptr2023}         & 3.7 & 10.5 & 10.6 & 8.3 \parnum{40.1} & 4.9 & 12.8 & 13.9 & 10.5 \parnum{39.9} & 4.3 & 13.9 & 14.8 & 11.0 \parnum{45.4} & 9.0 & 18.4 & 17.5 & 15.0 \parnum{53.1} \\
MapTRv2~\citep{maptrv22024}         & 4.3 & 3.9 & 4.3 & 4.2 \parnum{34.8} & 8.0 & 8.2 & 10.0 & 8.7 \parnum{38.8} & 5.2 & 7.5 & 8.4 & 7.1 \parnum{43.8} & 13.2 & 12.0 & 14.6 & 13.3 \parnum{54.7} \\
SMN~\citep{streammapnet}           & 2.3 & 10.1 & 11.1 & 7.9 \parnum{26.7} & 4.5 & 16.7 & 13.2 & 11.5 \parnum{33.3} & 4.5 & 20.9 & 17.9 & 14.4 \parnum{41.8} & 12.5 & 20.6 & 24.6 & 19.3 \parnum{55.0} \\
GeMap~\citep{gemap2024}         & 3.7 & 7.7 & 9.9 & 7.2 \parnum{41.2} & 8.0 & 12.2 & 16.2 & 12.2 \parnum{39.5} & 6.7 & 12.5 & 15.2 & 11.4 \parnum{48.3} & 12.7 & 14.0 & 15.7 & 14.1 \parnum{60.9} \\
MapQR~\citep{liu_leveraging_2024}        & 3.4 & 7.9 & 8.9 & 6.8 \parnum{32.3} & 6.1 & 12.7 & 12.8 & 10.5 \parnum{40.7} & 5.1 & 14.1 & 14.8 & 11.3 \parnum{49.5} & 11.6 & 19.2 & 20.8 & 17.2 \parnum{62.9} \\
\midrule
MapTRv2$^\ast$ & 22.0 & 21.5 & 21.6 & 21.7 & 18.5 & 19.0 & 22.2 & 19.9 & 20.0 & 24.5 & 26.9 & 23.8 & 30.5 & 28.0 & 33.9 & 30.8 \\
\textbf{Ours} & \textbf{38.2} & \textbf{49.7} & \textbf{48.0} & \textbf{45.3}
& \textbf{29.9} & \textbf{48.6} & \textbf{50.8} & \textbf{43.1}
& \textbf{33.6} & \textbf{49.8} & \textbf{48.9} & \textbf{44.1}
& \textbf{41.7} & \textbf{43.0} & \textbf{49.7} & \textbf{44.8} \\
\bottomrule
\end{tabular}%
}

\label{tab:quan_nusc}
\end{table*}

\subsection{Experimental setup}
\label{sec:exp_setup}

\noindent\textbf{Datasets.} 
We evaluate FlexMap on nuScenes~\citep{caesar_nuscenes_2020}, which contains 1,000 driving scenes with synchronized images from six cameras at 2 Hz. We use HD map annotations for three classes: lane dividers, pedestrian crossings, and road boundaries. We also evaluate on Argoverse 2~\citep{wilson_argoverse_2023} to test the method across datasets and camera setups. Argoverse 2 contains 1,000 driving logs, each with 15 seconds of RGB images from seven cameras at 20 Hz and 3D vectorized map annotations for the surrounding road geometry.

\noindent\textbf{Evaluation metrics.}
We follow standard vectorized map evaluation protocols~\citep{maptr2023,li2021hdmapnet,liu_vectormapnet_2023}, with one distinction: predictions are evaluated for each camera view instead of first being merged into a single BEV map. Average Precision (AP) measures detection quality, and Chamfer distance ($D_{\text{Chamfer}}$) measures geometric alignment between predicted and ground-truth polylines. For each map class, we compute $\text{AP}_{\mathcal{T}}$ at $\mathcal{T} \in \{0.5\text{m},1.0\text{m},1.5\text{m}\}$ and average across thresholds and classes to obtain mAP. The reported AP values are also averaged across input views. For scene-level evaluation, we merge per-view predictions and rasterize predictions and ground truth on a $200{\times}400$ pixel BEV canvas. Vectors are rasterized with a 2 m line width to form per-class binary masks. We average Intersection over Union across these masks to obtain scene-level mIoU.
\subsection{Implementation and evaluation protocol}
\begin{figure}[!ht]
    \centering
    \includegraphics[width=0.9\linewidth]{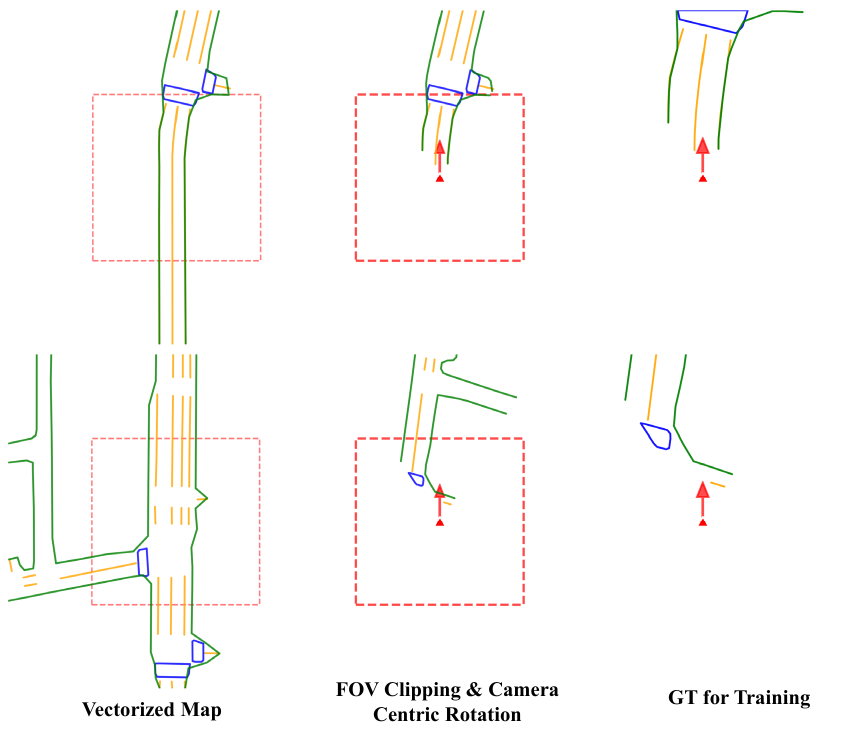}
    \caption{Illustration of our offline data processing pipeline. Ground truth map elements are extracted as a $120\times120$m BEV patch, clipped to the specific camera's FOV, rotated into a camera-centric coordinate frame, and finally cropped to the $60\times60$m target region used by the decoder for camera-centric prediction.}
    \label{fig:appendix_gt_process}
\end{figure}

\noindent\textbf{Data processing.}
We use a four-step ground-truth processing pipeline to produce consistent camera-centric map targets for all camera orientations. The process retains sufficient map coverage before converting each target to the coordinate frame used by the decoder, as illustrated in \cref{fig:appendix_gt_process}:

\noindent\textbf{Step 1: Large patch extraction (120×120m).} We first extract a 120×120m BEV map patch centered at the ego vehicle's LiDAR position. This patch is twice the size of the final 60×60m target, leaving enough map coverage for the later rotation, particularly for rear-facing cameras.

\noindent\textbf{Step 2: Camera FOV clipping.} We transform map points to 3D camera space and clip those outside the camera's horizontal field of view (FOV). The remaining vectors therefore correspond to map elements within the selected camera's FOV and define its camera-centric target.

\noindent\textbf{Step 3: Camera-centric rotation.} We rotate the FOV-clipped vectors into a camera-centric frame whose +Y axis follows the camera's forward direction. This transformation gives every camera view a common output convention and allows different camera orientations to be used in the same training batch under a shared output definition.

\begin{figure}[t]
    \centering
    \includegraphics[width=\linewidth]{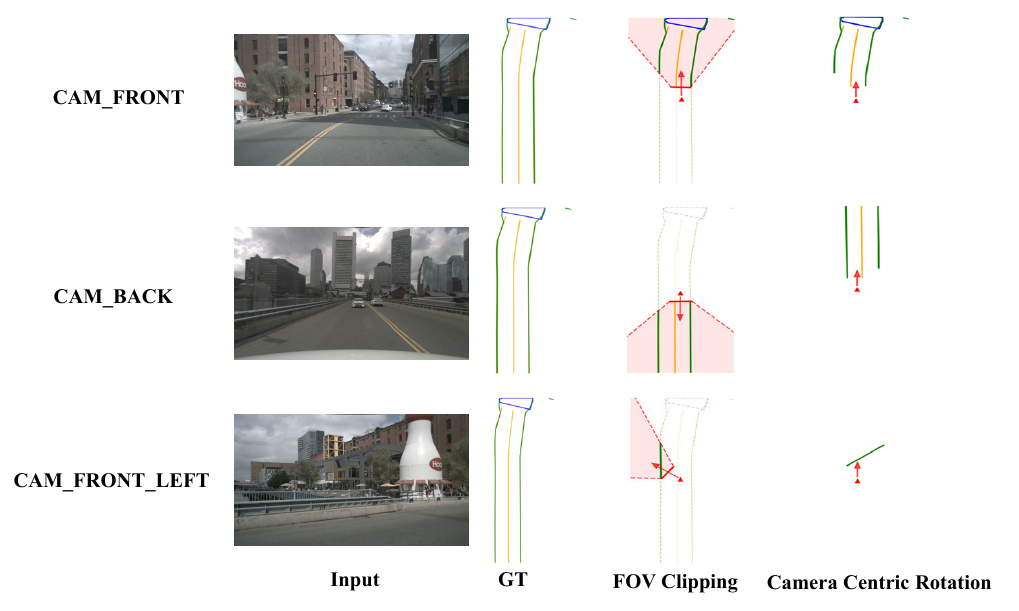}
    \caption{Illustration of the evaluation pipeline used to compare our method with BEV-based baselines. Baseline predictions are transformed from ego-centric BEV coordinates into camera-centric coordinates and then clipped to the visible field of view for fair per-camera evaluation across configurations.}

    \label{fig:appendix_eval_process}
\end{figure}

\noindent\textbf{Step 4: Final cropping (60×60m).} After rotation, we crop the vectors to the final 60×60m region. Unlike protocols that extract a 30×60m target region directly~\cite{maptr2023,gemap2024}, the larger initial patch prevents content from being lost when the vectors are rotated into camera-centric coordinates. This design supports joint training with all six camera views.
We apply the same FOV-based protocol to every method (\cref{fig:appendix_eval_process}). MapTR~\cite{maptr2023}, MapTRv2~\cite{maptrv22024}, StreamMapNet~\cite{streammapnet}, GeMap~\cite{gemap2024}, and MapQR~\cite{liu_leveraging_2024} predict maps in ego-centric BEV coordinates. We transform their outputs to the same camera-centric coordinates and apply the same FOV clipping used for FlexMap and the ground truth. Thus, every method is evaluated on the same FOV-defined region for each camera. Predictions are then matched to ground truth with the same matching cost and Chamfer thresholds $\tau \in \{0.5,1.0,1.5\}$m, following MapTR~\cite{maptr2023}.

\begin{table*}[!t]
\centering
\caption{Comparison on the Argoverse 2~\citep{wilson_argoverse_2023} validation set across camera configurations. Baselines use estimated poses, FlexMap uses no pose input, and all temporal methods use the same input length.}
\label{tab:argo_mono_baselines}
\setlength{\tabcolsep}{4.0pt}
\renewcommand{\arraystretch}{1.1}

\resizebox{0.8\textwidth}{!}{%
\begin{tabular}{lcccc|cccc|cccc}
\toprule
& \multicolumn{4}{c|}{\textbf{Front camera (1PV)}} 
& \multicolumn{4}{c|}{\textbf{Rear cameras (2PV)}}
& \multicolumn{4}{c}{\textbf{Full camera suite}} \\
\cmidrule(lr){2-5}\cmidrule(lr){6-9}\cmidrule(lr){10-13}
\textbf{Method}
& $\text{AP}_{\text{ped}}$ & $\text{AP}_{\text{div}}$ & $\text{AP}_{\text{bou}}$ & \textbf{mAP}
& $\text{AP}_{\text{ped}}$ & $\text{AP}_{\text{div}}$ & $\text{AP}_{\text{bou}}$ & \textbf{mAP}
& $\text{AP}_{\text{ped}}$ & $\text{AP}_{\text{div}}$ & $\text{AP}_{\text{bou}}$ & \textbf{mAP} \\
\midrule
MapTRv2~\citep{maptrv22024}  & 6.1 & 14.6 & 11.7 & 10.8 & 8.1 & 17.8 & 14.0 & 13.3 & 13.5 & 28.0 & 23.0 & 21.5 \\
MapQR~\citep{liu_leveraging_2024}    & 3.1 & 6.1 & 5.2 & 4.8 & 3.0 & 7.9 & 6.5 & 5.8 & 8.4 & 18.0 & 15.0 & 13.8 \\
\midrule
\textbf{Ours (FlexMap)} 
         & \textbf{38.5} & \textbf{48.2} & \textbf{49.5} & \textbf{45.4}
         & \textbf{33.6} & \textbf{52.5} & \textbf{52.2} & \textbf{46.1}
         & \textbf{36.8} & \textbf{52.5} & \textbf{53.5} & \textbf{47.6} \\
\bottomrule
\end{tabular}%
}
\end{table*}

\label{sec:appendix_eval}
\noindent\textbf{Training details.} We train our model with the VGGT~\citep{wang_vggt_2025} backbone frozen, using the AdamW optimizer with a learning rate of $2\times10^{-4}$ and weight decay 0.01. A cosine annealing scheduler reduces the learning rate to a minimum of $1\times10^{-5}$. Dynamic batch sampling adjusts batch size inversely with sequence length to maintain constant GPU memory usage, processing 48 frames per GPU during training. Sequence lengths are sampled from $S \in \{2,\ldots, 24\}$ frames, where $S$ represents the total number of camera views across timestamps. Gradient checkpointing is enabled to reduce memory consumption. The pyramid fusion module that extracts features from VGGT layers $\{4, 11, 17, 23\}$, the spatial-temporal enhancer with four attention layers and window size five, and the camera-aware decoder with six deformable attention layers are trained from scratch. Loss weights are set to $\lambda_{\text{cls}}=3.0$, $\lambda_{\text{pts}}=4.0$, and $\lambda_{\text{dir}}=0.05$ for the classification, point regression, and direction losses, respectively, in the joint training objective.

\subsection{Baselines}
\label{sec:metrics_baselines}

We compare FlexMap with five recent vectorized HD mapping methods: MapTR~\citep{maptr2023}, MapTRv2~\citep{maptrv22024}, StreamMapNet (SMN)~\citep{streammapnet}, GeMap~\citep{gemap2024}, and MapQR~\citep{liu_leveraging_2024}. These baselines cover hierarchical map queries, temporal mapping, geometric modeling, and enhanced query designs. For the standard baseline rows, we mask unavailable camera inputs at inference while keeping each model's pre-trained weights fixed; we also report a separately retrained MapTRv2 model. Every method, including FlexMap, uses the same FOV-based output clipping and per-view evaluation region for every camera configuration.

\begin{figure*}[ht]
    \centering
    \includegraphics[width=0.9\linewidth]{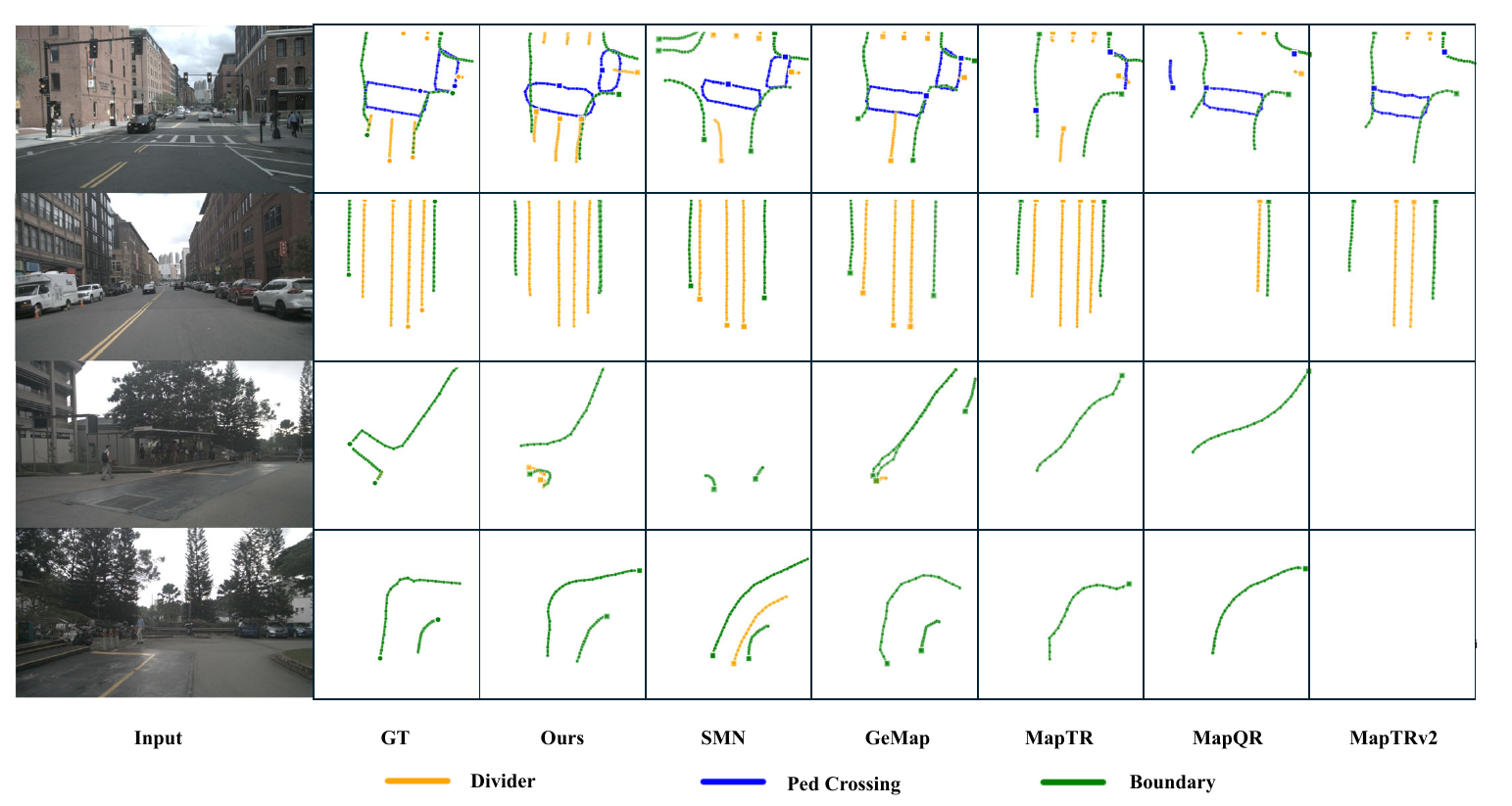}
    \caption{\textbf{Qualitative comparison on nuScenes dataset.} We visualize predictions from different methods across challenging scenarios using a single front camera as input. Without requiring camera pose as model input, FlexMap produces predictions (third column) that more closely follow the ground-truth geometry (second column) than baselines supplied with ground-truth camera poses.
    }
    \label{fig:qualitative}
\end{figure*}

\subsection{Quantitative results}
\label{sec:quantitative}
We evaluate baselines in two settings. The first supplies ground-truth camera poses and is reported in parentheses as a reference. The second replaces ground-truth poses with VGGT estimates~\citep{wang_vggt_2025}, while FlexMap receives no pose input. All temporal methods use the same temporal input length and therefore receive the same amount of temporal context. The estimated-pose setting is our main comparison because it evaluates every method without ground-truth pose input at inference.

\Cref{tab:quan_nusc} reports nuScenes results across four camera configurations. FlexMap achieves 45.3 mAP with the front camera, 43.1 with the back camera, 44.1 with front and back cameras, and 44.8 with all six cameras. It outperforms every baseline that uses estimated poses for both single-view and multi-view inputs. We also retrain MapTRv2 with estimated poses to align its training protocol with the estimated-pose evaluation. The retrained model reaches 30.8 mAP for 6PV, compared with 44.8 for FlexMap, and remains lower in all four configurations. This shows that one FlexMap model covers all four view sets without configuration-specific retraining or camera parameters as input.

\Cref{tab:argo_mono_baselines} reports Argoverse 2 results under the fixed temporal-input protocol, with estimated poses supplied to the baselines. This dataset uses a different seven-camera arrangement and captures different urban environments from nuScenes. FlexMap achieves 45.4 mAP with the front camera, 46.1 with the rear cameras, and 47.6 with the full camera suite. It exceeds MapTRv2~\citep{maptrv22024} and MapQR~\citep{liu_leveraging_2024} in all three configurations. The same ordering therefore holds on a different camera rig and dataset.

\subsection{Analysis across camera configurations}
The configuration-level results characterize how FlexMap behaves as view direction and camera count change. Its nuScenes results span 43.1--45.3 mAP across front-only, back-only, paired-view, and six-view inputs, while its Argoverse 2 results span 45.4--47.6 mAP on a different seven-camera rig. Because predictions and metrics are defined on camera-specific targets, these ranges indicate comparable per-view mapping quality as both the active view set and camera arrangement change.

The per-view evaluation explains the non-monotonic relation between mAP and camera count. Each configuration evaluates the camera-centric outputs of its active views, so adding cameras expands the set of view-specific targets rather than adding observations to one fixed BEV target. The class-wise results show how these targets vary across configurations. Boundary AP stays between 48.0 and 50.8, while pedestrian-crossing AP ranges from 29.9 to 41.7 as the evaluated views change. FlexMap uses the same camera-centric output definition for these different target distributions. Its view-conditioned queries and sampling locations adapt the decoder to each view, enabling one model to cover different view counts and orientations without constructing a configuration-specific BEV grid.

Retraining MapTRv2 with estimated poses provides an adapted baseline for the main comparison. The retrained model obtains 21.7, 19.9, 23.8, and 30.8 mAP across the four nuScenes configurations, with its strongest result in the six-view setting. Every method uses the same FOV-defined output region, and all temporal methods use the same input length. Under this protocol, FlexMap remains more accurate in all four configurations while decoding geometry-grounded, camera-centric features without alignment through estimated camera parameters.

The two approaches process changing camera inputs differently. Projection-based baselines place the features from each camera into a shared BEV frame using the supplied camera geometry, so the available BEV evidence changes with the active view set. FlexMap instead keeps dense features in perspective-view space and builds cross-view context inside the geometry transformer before vector decoding. Its decoder uses the camera token to initialize queries and select sampling locations for each view. Changing the active cameras changes the feature sequence, while the decoder and camera-centric output definition remain unchanged. This common decoding rule avoids switching models or constructing a configuration-specific BEV representation as the active view set changes.

\begin{figure}[t]
    \centering
    \includegraphics[width=0.95\linewidth]{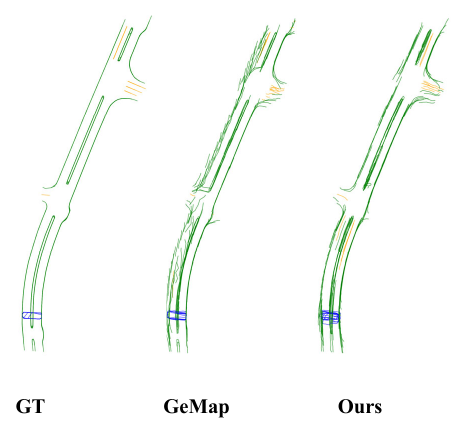}
    \caption{\textbf{Qualitative comparison on nuScenes~\cite{caesar_nuscenes_2020} with front-only input.}
    We visualize scene-level reconstructions from baselines and our method using a single front camera. All predictions are aligned using ground-truth poses for visualization, while FlexMap does not use camera pose for map prediction. Overall, our predictions exhibit cleaner map geometry, fewer fragmented dividers, and more consistent boundary connectivity.}

    \label{fig:qualitative_whole_scene}
\end{figure}
For scene-level evaluation, we align per-view predictions using ground-truth location and camera poses, then compute rasterization-based metrics. This tests scene-map consistency after camera-centric predictions are aligned. FlexMap and GeMap use the same alignment and rasterization procedure. With front-only input on the nuScenes validation set, FlexMap achieves 47.30\% mIoU, compared with 38.10\% for GeMap. The higher mIoU shows that the camera-centric predictions remain consistent after sequence-level assembly over complete driving sequences.

Per-view AP measures map-element accuracy for each camera-centric prediction, and scene-level mIoU measures consistency after alignment. The two metrics therefore evaluate the same output at different scales: individual camera views and complete reconstructed scenes. Camera poses are introduced only after prediction to align camera-centric outputs for scene-level measurement; the FlexMap inference pipeline does not use them. \Cref{fig:qualitative_whole_scene} complements the metric with visualizations of lane spacing, boundary continuity, and intersection structure across the assembled scene map under front-only input, linking per-view prediction quality to complete-scene consistency over full sequences.

\subsection{Qualitative analysis}
\label{sec:qualitative}

\begin{figure}[!tbp]
    \centering
    \includegraphics[width=0.95\linewidth]{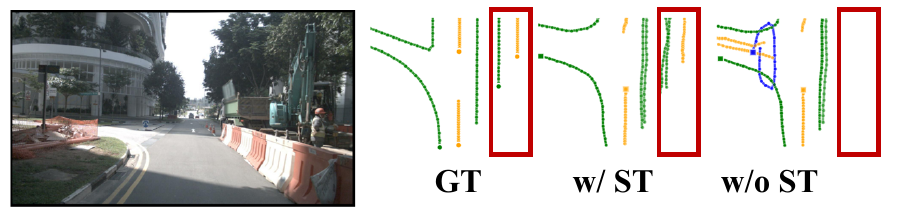}
    \caption{\textbf{Ablation of the spatial-temporal attention module on nuScenes.} We compare predictions with and without spatial-temporal enhancement against the ground truth. With the module, lane predictions remain more continuous around occluded regions.}
    \label{fig:spatio_temporal}
\end{figure}

All visualized baselines use ground-truth camera poses. \Cref{fig:qualitative} compares four challenging scenes. At the urban intersection (row 1), FlexMap detects all pedestrian crossings and follows the ground-truth lane dividers and road boundaries. MapQR and MapTRv2~\citep{maptrv22024} miss crossings, GeMap~\citep{gemap2024} adds false predictions, and MapTR~\citep{maptr2023} produces fragmented crossing boundaries. On the straight highway (row 2), FlexMap preserves clean lane geometry and consistent spacing over the visible range. On the curved road (row 3), FlexMap follows the road curvature and preserves lane continuity; the baselines produce more fragmented predictions. Under heavy tree occlusion (row 4), FlexMap preserves road structure through partially hidden regions, consistent with \cref{fig:spatio_temporal}, where spatial-temporal enhancement improves continuity around occluded regions.

\Cref{fig:qualitative_whole_scene} compares complete scene-level maps from FlexMap and GeMap~\citep{gemap2024}. FlexMap produces a cleaner and more coherent vector map over the full sequence. Parallel lanes remain separated over longer distances, intersections preserve their connectivity, and the overall spatial layout follows the ground truth more closely. The higher mIoU confirms that these local predictions assemble into a scene map with consistent global geometry after alignment.

\begin{table}[!htbp]
\centering
\caption{Ablation study on nuScenes~\citep{caesar_nuscenes_2020} validation set. A: spatial-temporal attention enhancement; B: camera-aware decoder. Metrics are computed over predictions from eight predefined camera configurations used in the ablation.}

\setlength{\tabcolsep}{6.0pt}
\renewcommand{\arraystretch}{1.1}
\resizebox{0.4\textwidth}{!}{%
\begin{tabular}{lcccccc}
\toprule
\textbf{Variant} & \textbf{A} & \textbf{B} &
$\text{AP}_{\text{ped}}$ & $\text{AP}_{\text{div}}$ & $\text{AP}_{\text{bou}}$ & \textbf{mAP} \\
\midrule
Baseline     &            &            & 30.8 & 46.9 & 48.8 & 42.2 \\
+ A          & \checkmark &            & 31.0 & 50.6 & 50.3 & 44.0 \\
+ B          &            & \checkmark & 34.2 & 50.8 & 52.0 & 45.7 \\
\midrule
\textbf{A + B} 
& \checkmark & \checkmark & 
\textbf{36.3} & \textbf{51.7} & \textbf{52.7} & \textbf{46.9} \\
\bottomrule
\end{tabular}
}

\label{tab:ablation_ab}
\end{table}
\subsection{Ablation study}
\label{sec:ablation}

\Cref{tab:ablation_ab} analyzes two components: (A) spatial-temporal enhancement and (B) the camera-aware decoder with camera tokens. Unlike \cref{tab:quan_nusc}, which reports four representative configurations individually, the ablation metrics are computed over predictions from eight predefined camera configurations. The baseline without either component obtains 42.2 mAP. Adding A, which separates cross-view spatial aggregation from cross-time temporal aggregation, increases mAP to 44.0. Adding B, which introduces view-conditioned query initialization and view-adaptive attention, increases mAP to 45.7. Combining both components reaches 46.9 mAP, exceeding the single-component variants by 2.9 and 1.2 mAP, respectively. The full model therefore benefits from both components. \Cref{fig:spatio_temporal} gives the corresponding qualitative comparison: the full model produces more continuous lanes around occluded regions.

\subsection{Computational efficiency and deployment}
\label{sec:efficiency}
Training uses eight NVIDIA A100 (80GB) GPUs and takes approximately three days. At inference time, FlexMap requires neither camera-pose input nor configuration-specific mapping models. On a single NVIDIA RTX 4090, it runs at ${\sim}$12 FPS for one perspective view (1PV) and ${\sim}$2 FPS for all six surrounding views (6PV). Runtime varies with the number of processed views, while the decoder structure and output definition remain unchanged. Thus, one model supports different camera configurations, with computation scaling according to the number of active views.

\section{Conclusion}

We presented FlexMap, a framework for vectorized HD map construction across flexible camera configurations. FlexMap replaces explicit 2D-to-BEV projection with a geometry foundation model, allowing the same architecture to process single-camera and multi-camera inputs without camera parameters as model input. Its spatial-temporal enhancement module separates cross-view aggregation from temporal modeling, and the camera-aware decoder adapts queries and attention to each input view. Across front-only, rear-only, partial-surround, and full-surround inputs, FlexMap outperforms baselines that use estimated poses on nuScenes and Argoverse 2. The decoder and camera-centric output definition remain unchanged across these settings. The resulting predictions also form coherent scene maps when assembled over complete sequences. FlexMap thus supports HD map construction across heterogeneous camera configurations with one shared model.

{
    \small
    \bibliographystyle{ieeenat_fullname}
    \bibliography{references}
}

\end{document}